\ifdefined\pdfoutput\pdfoutput=1\fi
\documentclass[acmsmall,screen,nonacm]{acmart}

\usepackage{booktabs}
\usepackage{tabularx}
\usepackage{array}
\usepackage{graphicx}
\newsavebox{\tblbox}  
\usepackage{tikz}
\usetikzlibrary{arrows.meta,positioning,shapes.geometric,fit,backgrounds}
\renewcommand\footnotetextcopyrightpermission[1]{}
\setcopyright{none}
\acmYear{2026}
\acmDOI{}
\acmISBN{}

\newcommand{\ARTIFACTURL}{archival deposit with a persistent identifier to follow the peer-reviewed version, available from the corresponding author meanwhile}
\newcommand{\OSFURL}{\url{https://osf.io/enjha/?view_only=40f7fc5fdd0043e0ad052b1d3017b209}}

\begin{document}

\title{Harness or Model? Isolating the Harness Effect in Agentic Coding with a Contamination-Controlled Private Suite}
\titlenote{Revised 8 September 2026. A post-study review found a usage-semantics defect in the study's cost telemetry. Relative to the August 2026 manuscript, the cost results, the probe interpretation and the registration statement are corrected, and the capability analysis is re-run with the protocol's tests. The revision memo and reanalysis code are in the replication package.}

\author{Mohsen Arjmandi}
\orcid{0009-0006-0514-9244}
\email{mohsen.arjmandi@gmail.com}
\affiliation{%
  \institution{evolutionID GmbH}
  \country{Germany}
}

\begin{abstract}
An agentic coding system couples a language model to a harness: the tools, prompts, truncation policies and control flow that turn a chat model into an autonomous software engineer. Vendors ship harnesses tuned to their own models, and practitioners commonly assume that the vendor-native pairing solves more tasks. We measure that assumption directly with paired same-model contrasts on a private, contamination-controlled suite of 256 repository and post-cutoff contest tasks. The same 80 tasks were run under claude-agent-sdk and under deepagents (LangGraph) on claude-opus-4-8, and under the openai-codex SDK and deepagents on gpt-5.5, with gemini-3.5-flash and deepseek-v3.2 as side cells. Every run executed in its own KVM microVM with an append-only event ledger, and 792 of 800 planned runs were graded by a Docker-isolated oracle.

Neither contrast resolves an average advantage for either harness. The paired difference in solve rate is $-$1.25 pp for Opus 4.8 (48.8\% native against 50.0\% neutral, task-bootstrap 95\% CI [$-$10.0, +7.5]) and +1.25 pp for GPT-5.5 (55.6\% against 54.4\%, CI [$-$4.4, +6.9]), conditional on the selected pool. The Opus average combines opposite strata: the native harness trails by 9.0 pp on the 61 repository tasks (CI [$-$17.2, $-$0.8]) and leads by 23.7 pp on the 19 contest tasks (CI [+2.6, +44.7]). A permutation test of the workload labels gives p = 0.003 for this interaction. The partition was chosen after seeing the data, so we report it as a post-hoc pattern that a designed replication should test. Correctness and autonomous completion also separate: 22 of 81 runs that were cancelled at the wall-clock ceiling had produced a passing patch. Re-priced from the raw per-turn usage at frozen list prices, the neutral harness cost 1.3–1.6 times as much per solved task on Opus 4.8 (task-bootstrap intervals 1.1–2.1 across pricing bases) and 1.2 times on GPT-5.5 (1.05–1.33). These are observed-usage estimates. On the Anthropic account 58 runs ended without a usage record, and allocating that unrecorded spend entirely to either cell would move the Opus ratio between 0.7 and 2.3, so the billed ordering there is unresolved. A twelve-task probe of one truncation setting was inconclusive. This revision corrects an August 2026 manuscript whose cost figures rested on a usage-semantics defect in our own telemetry, described in \S5.1 and itemised in the revision record. We release the orchestrator, grading oracle, reanalysis code and derived aggregates. The tasks stay private to preserve the suite.

\end{abstract}



\maketitle

\section{Introduction}

Benchmarks of coding agents usually vary the model and hold the harness fixed, or they report a vendor's harness and model together as one opaque system. Neither design answers the question a team faces when it already has a frontier model and must choose the software that drives it. A vendor-native harness ties the deployment to one provider. A neutral harness such as deepagents promises portability across providers, at a capability and cost penalty that nobody has measured on private work.

We measure it with paired same-model contrasts. The model is held fixed and only the harness changes: claude-opus-4-8 under claude-agent-sdk and under deepagents (cells C1 and C2), and gpt-5.5 under the official openai-codex SDK and under deepagents (cells C3 and C4). Two side cells run gemini-3.5-flash and deepseek-v3.2 on the neutral harness. All six cells share one execution substrate: one microVM per run, identical repository seeds and prompts, a shared wire protocol, and a canonical append-only event history. The harness is the intended moving part within each contrast. \S7 lists the configuration differences that remain, such as the smaller VM memory of the codex cell.

The study was planned in June 2026 and executed in August 2026 around two questions fixed in the program plan before any scored run. RQ1 asks whether a vendor-native harness has a capability premium over a neutral harness on the same model. The plan's hypotheses H1 to H3 cover existence, vendor dependence and concentration by task type. RQ2 asks whether the capability winner is also the cost-per-solved-task winner (H4).

Two methodological problems shape any such study. The first is contamination. Public benchmarks leak into training corpora, so our suite is private: 179 repository tasks mined from four production codebases (AccessCtl, IdentityApp, FieldSvc, BookingSvc) with hidden tests and gold patches, plus 77 contest tasks published after a mechanically derived eligibility date. A machine-readable cutoff registry was frozen before data collection, and a runtime drift gate verifies the served model identity before every scored phase. The tasks stay private, and we release the methodology, the registry format and the orchestration code.

The second problem is cost measurement, and here the study became its own case. Agent evaluations increasingly report dollar costs derived from harness-reported token usage. Our telemetry normalizer applied one SDK's convention for cache tokens to all three harnesses, and the August 2026 version of this paper built its cost conclusions on the result. A post-study review found the defect (\S5.1). This revision re-derives every cost figure from the raw per-turn events and records every changed claim in the revision record that accompanies the replication package.

\textbf{Contributions.}

\begin{enumerate}
  \item Paired same-model harness contrasts on a private, contamination-controlled suite (80 tasks per contrast, 792 graded runs), reported with task-level bootstrap intervals and the tests specified in the protocol.
  \item A post-hoc workload pattern on Opus 4.8 in which the sign of the harness effect differs between repository and contest tasks, tested with an explicit interaction null and shown to combine arithmetically into the overall estimate.
  \item A separation of two endpoints, correct output and autonomous completion, using wall-clock ceilings, terminal states and latency.
  \item A usage-semantics defect and its correction as a methods result: input-token fields are cache-inclusive on some SDKs and cache-exclusive on others, and a normalizer that assumes one convention doubles the other. We give a reproducible source-to-ledger reconciliation and an account-level check against billed spend.
  \item A runtime model-drift gate with two live catches, and release of the harness-neutral orchestrator, grading oracle, reanalysis code and derived aggregates.
\end{enumerate}

\section{Benchmark Design}

\subsection{Task suite}

The suite has \textbf{256 tasks} in two tracks. \textbf{Track A (179 repository tasks)} was mined from four private production codebases owned by our organization: AccessCtl (110 tasks), IdentityApp (27), FieldSvc (40) and BookingSvc (2). Each task follows the SWE-bench pattern of FAIL\_TO\_PASS and PASS\_TO\_PASS hidden tests plus a gold patch. \textbf{Track B (77 contest tasks)} consists of LeetCode weekly and AtCoder problems published strictly after the eligibility date, each with a hidden test suite (2.0 GB in aggregate).

Task prompts were finalized under a documented author and reviewer protocol. 142 mined prompts were rewritten and 37 hand-authored, and all 179 Track A prompts were reviewed to a \texttt{reviewer-final-v1 (no-spoiler)} standard by an AI author, an AI reviewer and human spot-checks. 51 of 179 required fixes. One suite property is known but not measurable from the archive. Roughly 30 Track A tasks import internal symbols by exact name, which we call white-box coupling, and this depresses solve rates independently of intrinsic difficulty. The per-task labels were not retained, so the planned sensitivity analysis cannot be run and we cannot tell whether the coupling favours one harness (\S7).

Every grading environment is manifest-driven Docker with service sidecars where the codebase requires them (PostgreSQL for AccessCtl, MSSQL for FieldSvc, a mock-OIDC service for BookingSvc). Each environment was validated before any scored run: the gold patch must pass and the base tree must fail, with determinism enforced by oracle invariants. This pre-flight caught and fixed one defective gold patch during qualification.

\subsection{Contamination protocol}

The suite is private and post-cutoff by construction, with four defence layers.

\textbf{Frozen cutoff registry and eligibility arithmetic.} We froze a machine-readable registry of model training cutoffs (\texttt{eval/\allowbreak{}plans/\allowbreak{}cutoffs.frozen.md} + \texttt{.json}) on 2026-06-24, before any scored run. The eligibility date, \textbf{2026-03-02}, is the newest binding cutoff among study models (Opus 4.8, 2026-01-31) plus a 30-day buffer. All Track B problems post-date it, and Track A mirrors follow an ancestors-of-base leakage rule. The registry also documents look-alike hazards. gemini-3.5-flash-lite, for example, carries a March 2026 cutoff past our eligibility date, so a silent model substitution would void the control.

\textbf{Runtime drift gate.} Every run leg begins with a pre-flight gate (\texttt{eval/\allowbreak{}orchestrator/\allowbreak{}preflight.py}) that verifies the served model identifier, with dated-snapshot normalization, and soft-checks the model's self-reported cutoff. It correctly normalized OpenAI serving \texttt{gpt-5.5-2026-04-23} for the pinned \texttt{gpt-5.5}, and it observed DeepSeek's self-reported cutoff changing across probes (2023-07, 2023-10, 2024-07). Self-report is therefore advisory only. After an incident in which the gate probed OpenRouter's default route while the study ran on a DeepInfra-pinned route, the probe was amended to test the run's exact pinned backend (A7).

\textbf{Canaries, mirror hygiene, and the scope of the claim.} Canary strings are embedded in private task content. The mining repository was never pushed, agent-facing run mirrors were private, and gold patches, hidden tests and canaries never entered a mirror. The mirrors were deleted after the study (\S7). The registry and the drift gate rule out documented training exposure to the contest problems, and the repository tasks are private by construction. They do not rule out exposure through channels a cutoff registry cannot see, and we did not run the canary emission probe on the study models. The claim is therefore bounded to what documented cutoffs and private provenance can establish.

\subsection{Experimental matrix, protocol deposit, and amendments}

\textbf{The 6-cell matrix} crosses three harnesses with four models:

\begin{table}[t]
\centering\footnotesize\setlength{\tabcolsep}{3.5pt}
\caption{Cells}
\label{tab:2-1}
\begin{tabularx}{\linewidth}{>{\hsize=0.34\hsize\raggedright\arraybackslash}X>{\hsize=1.93\hsize\raggedright\arraybackslash}X>{\hsize=0.98\hsize\raggedright\arraybackslash}X>{\hsize=0.83\hsize\raggedright\arraybackslash}X>{\hsize=0.91\hsize\raggedright\arraybackslash}X}
\toprule
Cell & Harness & Provider & Model & Role \\
\midrule
C1 & claude-sdk (claude-agent-sdk 0.2.130) & anthropic & claude-opus-4-8 & vendor-native \\
C2 & deepagent-sdk (deepagents 0.7.4 / LangGraph 1.2.10) & anthropic & claude-opus-4-8 & neutral, same model \\
C3 & codex-sdk (openai-codex 0.1.0b3, CLI 0.137.0a4) & openai & gpt-5.5 & vendor-native \\
C4 & deepagent-sdk & openai & gpt-5.5 & neutral, same model \\
C5 & deepagent-sdk & google & gemini-3.5-flash & cheap-frontier side cell \\
C6 & deepagent-sdk & openrouter (DeepInfra pin) & deepseek/deepseek-v3.2 & open-weights side cell \\
\bottomrule
\end{tabularx}
\end{table}

C1 against C2 and C3 against C4 hold the model fixed and vary only the harness (RQ1). C6's OpenRouter backend was pinned to DeepInfra with fallbacks disallowed, so that a reproducible evaluation never receives a different served quantization run-to-run. Library versions are those recorded on every session row of the archive.

\textbf{Solve metric and outcome taxonomy.} Each run's patch is extracted host-side (\texttt{git diff}) and graded offline by the Docker oracle with verdicts in \{pass, fail, timeout, apply\_error\}. A task's rate is the mean over repeats and a cell's rate the mean over tasks. Run outcomes are \{completed, ceiling\_wallclock, agent\_error, infra\}. Infra rows are never graded and are retried at most twice. The wall-clock ceiling (1,200 s) cancels the agent, and the oracle then grades whatever patch is present at cancellation. A ceilinged run can therefore pass (\S4.4). A repeat is an independent re-run of the same prompt, since no harness exposes a sampling seed.

\textbf{Phases and pool freeze.} The study ran in four phases. A dress rehearsal covered 14 tasks $\times$ 6 cells. A screening pass ran all 256 tasks under C1 and C3 at k = 1 (512/512 keys graded, C1 54.3\%, C3 58.2\%, agreement on 91\% of tasks). A deterministic pool freeze followed, and the main run covered the frozen pool. The selection rule is mechanical. All 24 tasks on which C1 and C3 disagreed in screening form the band. Remaining slots fill alternately from the both-pass and both-fail pools, ordered by sha256(task\_id), to a target of 80 (28 + 28). The frozen pool (contest 19, AccessCtl 36, IdentityApp 13, FieldSvc 12) is locked by hash (\texttt{final-80.lock.json}, sha256 \texttt{99cf4e42\ldots{}c575c}). Repeats were 2 for C1–C4 and 1 for C5 and C6, giving 800 main-run keys.

\textbf{What the selection rule does to the estimand.} Discordant tasks make up 9.4\% of the screening pool (24 of 256) and 30\% of the main pool (24 of 80). All main-run averages are therefore conditional on a pool that oversamples tasks on which the two vendor-native bundles disagreed. The pool is not a random sample of the suite, and the screening-blended sensitivity of \S4.2 reuses the selection data rather than replicating it.

\textbf{Planned power.} The program plan stated that the design could detect average premiums of roughly 8–10 pp and not smaller ones. Amendment A3 reduced the core cells from three to two repeats and moved the planned minimum detectable effect to about 12 pp. These are planning quantities under the plan's assumptions. The observed uncertainty is the task-bootstrap interval reported with each estimate in \S4.

\textbf{Protocol deposit.} The August 2026 manuscript described the design as pre-registered on OSF. The record is weaker, and we state it precisely. The program plan containing hypotheses H1 to H4, the cell design and the statistical plan (McNemar with cluster-bootstrap intervals, a mixed-effects logistic model for the nativeness $\times$ vendor interaction, BH-FDR, pass@k, and the power statement above) was first committed to the project repository on 2026-06-11, before any scored run. It is dated by version control, not by a registry. A private OSF project holds timestamped design documents deposited 25–28 June 2026: decision rules, the cutoff registry, canary and licensing notes, and a \texttt{PREREGISTRATION.md} (28 June) that describes a four-model benchmark comparison rather than the harness contrasts. No OSF registration was created, and the project is private. Its file listing (\OSFURL{}) shows the deposited files and their server dates. Table~\ref{tab:2-2} maps the plan to what was executed.

\begin{table}[t]
\centering\footnotesize\setlength{\tabcolsep}{3.5pt}
\caption{From protocol to execution}
\label{tab:2-2}
\begin{tabularx}{\linewidth}{>{\hsize=0.31\hsize\raggedright\arraybackslash}X>{\hsize=1.23\hsize\raggedright\arraybackslash}X>{\hsize=1.23\hsize\raggedright\arraybackslash}X>{\hsize=1.23\hsize\raggedright\arraybackslash}X}
\toprule
Element & Plan (repo, 2026-06-11) / OSF deposit (2026-06-28) & Executed (Aug 2026) & This revision \\
\midrule
Models & Sonnet 4.6, GPT-5.2(-Codex), Gemini 3.5 Flash, open-weight via groq/together & Opus 4.8, GPT-5.5, Gemini 3.5 Flash, DeepSeek-V3.2 via OpenRouter/DeepInfra & as executed \\
Repeats & k = 3 core cells & k = 2 core cells (A3) & as executed \\
Pool & $\approx$120 discrimination-screened tasks. 25–75\% band (OSF) & 80 tasks: 24 discordant + 28 + 28 by hash (A4) & as executed \\
H1 test & McNemar + cluster-bootstrap CI & n/a & task bootstrap, sign-flip, run-pair McNemar (\S4.2) \\
H2 test & mixed-effects logistic, nativeness $\times$ vendor & n/a & GEE logistic, task-clustered (\S4.2) \\
H3 test & premium by task type & n/a & workload split with interaction test, labelled exploratory (\S4.3) \\
Multiplicity & BH-FDR across pairwise claims & n/a & BH-FDR over seven declared tests (\S4.3) \\
pass@k & planned & n/a & pass@1, pass@2 (\S4.2) \\
Ceiling scoring & timeout = fail (OSF) & grader scores the patch at cancellation & as executed, disclosed (\S4.4) \\
Cost basis & raw tokens $\times$ frozen pricing table, SDK cost as cross-check & ledger from host telemetry & raw per-turn usage re-priced, ledger defect disclosed (\S5) \\
\bottomrule
\end{tabularx}
\end{table}

\textbf{Amendments.} Where operational reality forced changes, we amended before the affected data was collected and logged every amendment: A1 raised the C1 budget cap from \$4 to \$8 (the cap bound on FieldSvc in the dress rehearsal), A2 recalibrated the screening stop-loss, A3 reduced C2/C4 repeats from 3 to 2, A4 reduced the pool from 120 to 80 tasks with the rule retained, A5 gave the deepagent cells and later C1 8 GB sandboxes after out-of-memory runner deaths, A6 raised the operational main stop-loss for the finish leg, and A7 pinned the drift probe to the run's backend. None touches the scoring rule, the pool rule or the paired design.

\section{System and Execution}

\subsection{The platform as instrument}

\begin{figure}[t]
\centering
\begin{tikzpicture}[
  font=\footnotesize,
  node distance=4mm and 5mm,
  box/.style={draw, rounded corners=1.5pt, align=center, inner sep=3pt, minimum height=8mm, text width=21mm},
  store/.style={draw, cylinder, shape border rotate=90, aspect=0.15, align=center, inner sep=2pt, minimum height=8mm, text width=19mm},
  arr/.style={-{Latex[length=2mm]}, thick},
]
  \node[box] (plan) {run plan\\(YAML)};
  \node[box, right=of plan] (gate) {drift gate\\served-model check};
  \node[box, right=of gate] (sched) {scheduler\\conc.\,12, stop-loss,\\predictive brake};
  \node[box, right=of sched] (fact) {session factory\\deterministic ids,\\provenance\,=\,eval};
  \node[store, below=of sched] (ledger) {append-only\\ledger.jsonl};
  \begin{scope}[on background layer]
    \node[draw, dashed, rounded corners=2pt, fit=(plan)(gate)(sched)(fact)(ledger), inner sep=4pt,
          label={[font=\scriptsize\itshape, anchor=south west]north west:orchestrator (evaluation host)}] (orch) {};
  \end{scope}
  \node[box, right=of fact, text width=22mm] (rt) {agent runtime\\WebSocket driver,\\1200\,s ceiling};
  \node[box, below=of rt, text width=22mm] (vm) {per-run KVM microVM\\repo clone + SDK runner};
  \node[store, below=of vm] (trace) {trace.jsonl\\+ patch.diff};
  \node[box, left=of trace, text width=22mm] (grader) {grading oracle\\Docker, manifest\\sidecars, offline};
  \node[store, left=of grader] (verd) {verdicts.jsonl};
  \draw[arr] (plan) -- (gate);
  \draw[arr] (gate) -- (sched);
  \draw[arr] (sched) -- (fact);
  \draw[arr] (sched) -- (ledger);
  \draw[arr] (fact) -- (rt);
  \draw[arr] (rt) -- (vm);
  \draw[arr] (vm) -- node[right, font=\scriptsize] {wire msgs} (trace);
  \draw[arr] (trace) -- (grader);
  \draw[arr] (grader) -- (verd);
\end{tikzpicture}
\caption{One run is one microVM. The orchestrator drives sessions over the platform's WebSocket protocol, and grading is offline and post-hoc. Per-turn usage travels from the runner through the host normalizer to the session row and the ledger (\S5.1).}
\label{fig:arch}
\end{figure}
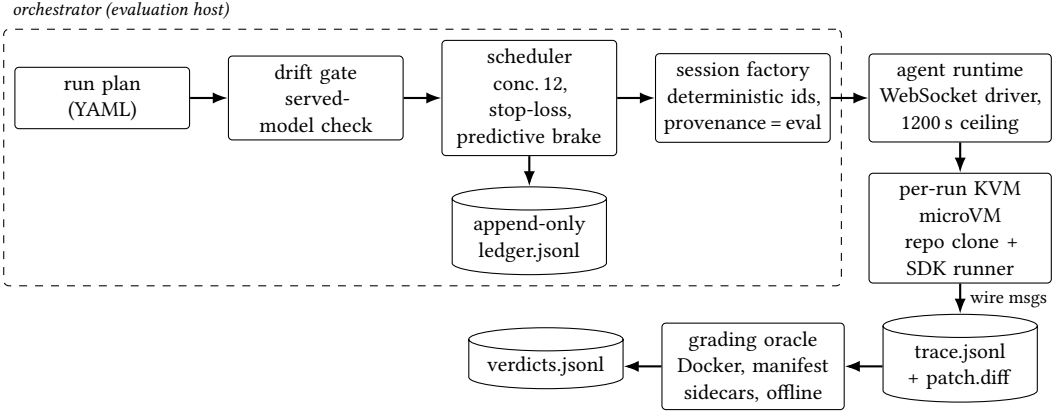

Every evaluation ran on our production multi-harness agent platform rather than on a purpose-built script. Each run is one chat session bound to one KVM microVM (microsandbox 0.3.14) holding a fresh clone of the task repository. The agent, one of claude-agent-sdk, the openai-codex SDK, or deepagents on LangGraph, executes as a runner subprocess inside that VM. Hypervisor isolation is the only sandbox, and the vendor harnesses' own approval and sandboxing layers are disabled, so all three face identical filesystem, network and process constraints. Every message the runner emits is persisted to an append-only canonical event log, which makes runs replayable and resumable. Figure~\ref{fig:arch} sketches the path.

\subsection{Orchestration, ledger, and telemetry path}

The orchestrator expands a run-plan YAML into a key space of (task $\times$ cell $\times$ repeat) runs and schedules them at concurrency 12. Each run's chat session is created with a deterministic id, \texttt{provenance = 'eval'}, pinned configuration and no resume token. A crash-resumed attempt reuses its session and a fresh attempt gets a fresh one, so retries never inherit poisoned history. The effective per-cell configuration is folded into a \texttt{config\_\allowbreak{}hash} on every ledger row. Every terminal run appends one JSON row (tokens, accounted cost, \texttt{end\_\allowbreak{}reason}, attempt) to \texttt{ledger.jsonl}, and \texttt{--resume} re-derives the remaining keys from it. A driver crash becomes a visible, retryable infra row, and an unrecordable ledger append halts the run, because spend the stop-loss cannot see must not be made. Each run has a hard 1,200 s wall-clock ceiling enforced in the WebSocket driver, and a per-run trace writer records every wire message alongside the extracted patch.

Spend control is an accounted-euro stop-loss over the ledger, priced from the host telemetry with a versioned pricing table, with a predictive brake that projects committed in-flight spend before each dispatch. Only infra rows are retried (at most twice), and graded outcomes are terminal. Failures caused by the experimenter's account rather than the model, such as provider credit exhaustion, are relabelled as retryable infra by a scrub step and re-run as fresh attempt-2 sessions. Across the main run and its finish leg the scrub recovered 71 poisoned rows. The archived main ledger holds 891 run rows for the 800 keys (800 first attempts, 70 second, 21 third) plus two halt records, and 8 keys remained infra-terminal. Relabelled rows keep their token usage, and \S5.3 accounts for them. The accounted euros were the operational currency for these controls. \S5 shows that they also overstated the deepagents cells with usable telemetry by roughly 5–8$\times$ and the Anthropic cells by a further 14\%. Appendix A summarises the incidents behind these mechanisms.

\subsection{The grading oracle}

Grading is offline. The orchestrator writes \texttt{grade: "ungraded"} on every ledger row, and a separate Docker-isolated oracle grades extracted patches against hidden tests with manifest-driven service sidecars per task family. Two invariants gate every environment: polarity (gold passes, base fails) and determinism (repeated grading agrees). The first full grading pass produced 85 grader errors that traced to a full disk. An idempotent purge-and-regrade of the 114 suspect keys yielded zero grader errors and 792 verdicts.

\subsection{Use of AI tools in this research}

AI tools were used as research instruments in three ways, each under human direction. \emph{Implementation and operation:} an LLM-based coding agent, operating on the platform under study, implemented the evaluation orchestrator and the grading oracle to the principal investigator's specification, executed the dress, screening, main and probe phases, and diagnosed the incidents of Appendix A. The PI made every configuration and go/no-go decision, including the prompt sign-off protocol, the repeat and pool reductions after a detectable-effect briefing, the stop-loss halt and finish-leg sequencing, and provider funding–. \emph{Task preparation:} task prompts were rewritten or authored by an LLM and reviewed by a second LLM pass to a no-spoiler standard, with human spot-checks, and 51 of 179 prompts were corrected (\S2.1). \emph{Analysis:} the statistics of \S4 and \S5 are computed by the released reanalysis script, written with LLM assistance. It reads only the checksummed archive and records the input digests, and every figure and table is regenerated from its output with assertions against the canonical results sheet. The agents evaluated in the study are the object of measurement, not instruments, and no AI tool grades patches. Grading is the deterministic oracle of \S3.3.

\section{Capability Results}

\subsection{What was graded, under which conventions}

All results come from the completed 800-key main matrix, the initial run halted by the operational stop-loss plus the finish leg that closed every remaining key. Table~\ref{tab:4-4} gives the full outcome flow. The grading oracle produced \textbf{792 verdicts: 380 pass, 404 fail, 7 timeout, 1 apply\_error}, with zero grader errors. Eight planned keys (C1: 1, C6: 7) ended infra-terminal after retries and were never graded.

Our published solve rate is the task-mean over pass/fail verdicts, which excludes timeout and apply\_error verdicts from a task's denominator. C5 and C6 therefore report 78 and 71 tasks, because two C5 tasks and two C6 tasks have only timeout verdicts at k = 1. Table~\ref{tab:4-1} reports two further conventions. Treating every non-pass verdict as a failure shifts the cell means by 0 pp for C1 and C2, $-$1.25 pp for C3, $-$0.6 pp for C4, $-$1.1 pp for C5 (denominator 80 tasks) and $-$0.5 pp for C6 (73 tasks). Counting non-pass over planned keys additionally counts the 8 infra-terminal keys and lowers C6 to 17.5\%. The paired Opus difference is identical under the first two conventions, and the GPT-5.5 difference moves from +1.25 to +0.6 pp.

\begin{table}[t]
\centering\small
\caption{Cell solve rates (frozen pool of 80 tasks)}
\label{tab:4-1}
\footnotesize\setlength{\tabcolsep}{4pt}
\sbox{\tblbox}{%
\begin{tabular}{llrrrr}
\toprule
Cell & Harness / model & Tasks & Solve rate & 95\% CI (pp) & All graded, non-pass $=$ fail (tasks) \\
\midrule
C1 & claude-sdk / Opus 4.8 & 80 & 48.8\% & $\pm$10.2 & 48.8\% (80) \\
C2 & deepagents / Opus 4.8 & 80 & 50.0\% & $\pm$10.3 & 50.0\% (80) \\
C3 & codex-sdk / GPT-5.5 & 80 & 55.6\% & $\pm$10.2 & 54.4\% (80) \\
C4 & deepagents / GPT-5.5 & 80 & 54.4\% & $\pm$10.5 & 53.8\% (80) \\
C5 & deepagents / Gemini 3.5 Flash & 78 & 44.9\% & $\pm$11.1 & 43.8\% (80) \\
C6 & deepagents / DeepSeek V3.2 & 71 & 19.7\% & $\pm$9.3 & 19.2\% (73) \\
\bottomrule
\end{tabular}}%
\ifdim\wd\tblbox>\linewidth\resizebox{\linewidth}{!}{\usebox{\tblbox}}\else\usebox{\tblbox}\fi

\end{table}

The interim solve rates computed at the stop-loss halt (454 verdicts) differed from the final ones. They were roughly 5 pp higher for the four frontier cells and lower for C6 (10.3\% interim against 19.7\% final). This pattern is consistent with the halt censoring slow runs, although two snapshots cannot establish the mechanism. Only the full-power figures are citable.

\subsection{The paired contrasts}

\begin{figure}[t]
\centering
\includegraphics[width=\linewidth]{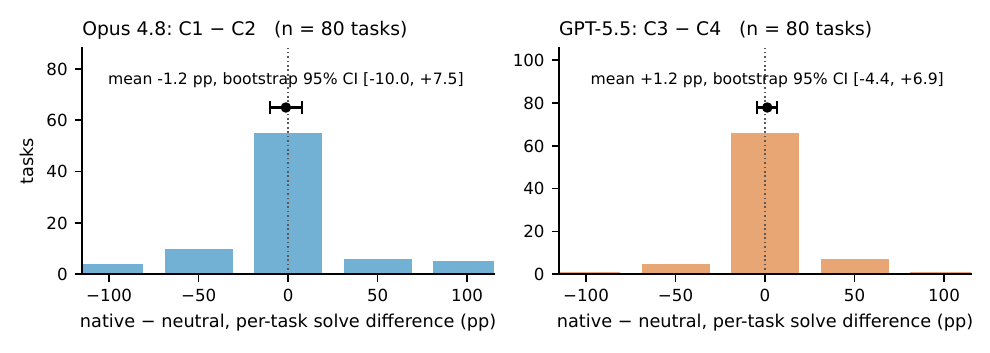}
\caption{Per-task paired outcomes for the two same-model contrasts (native $-$ neutral solve difference per task, $n=80$ each), with the mean paired difference and its task-bootstrap 95\% confidence interval. Neither estimate is distinguishable from zero, and both intervals admit differences of several points in either direction.}
\label{fig:paired}
\end{figure}

RQ1 asks whether a vendor's own harness outperforms a neutral harness driving the same model on the same tasks. Table~\ref{tab:4-2} reports the two paired contrasts. The primary uncertainty statement is the task-level bootstrap interval (10,000 resamples of tasks), because tasks are the sampling unit and repeats within a task are correlated. The normal-approximation intervals the August manuscript used, [$-$10.1, +7.6] and [$-$4.3, +6.8], are close to the bootstrap ones and are not repeated in the table. We declare one primary test per hypothesis (Appendix C of the preprint lists them all). For H1 the primary test is the sign-flip permutation on the per-task differences. The exact McNemar test on (task, repeat) run pairs pools correlated repeats and is supplementary. The cluster-adjusted McNemar statistic of Durkalski et al., which sums signed discordant counts within tasks, corrects for that pooling and is also reported. Figure~\ref{fig:paired} shows the per-task differences.

\begin{table}[t]
\centering\small
\caption{Paired native $-$ neutral contrasts (n = 80 tasks each)}
\label{tab:4-2}
\footnotesize\setlength{\tabcolsep}{4pt}
\sbox{\tblbox}{%
\begin{tabular}{llrlrlr}
\toprule
Contrast & Model & Difference & Task-bootstrap 95\% CI & Sign-flip $p$ & McNemar discordant ($p$) & Clustered McNemar $p$ \\
\midrule
C1 $-$ C2 & claude-opus-4-8 & \textbf{-1.25 pp} & [-10.0, +7.5] & 0.89 & 17 : 19 (0.87) & 0.78 \\
C3 $-$ C4 & gpt-5.5 & \textbf{+1.25 pp} & [-4.4, +6.9] & 0.83 & 11 : 9 (0.82) & 0.65 \\
\bottomrule
\end{tabular}}%
\ifdim\wd\tblbox>\linewidth\resizebox{\linewidth}{!}{\usebox{\tblbox}}\else\usebox{\tblbox}\fi

\end{table}

Neither contrast resolves an average advantage for either harness on this pool. The Opus interval runs from $-$10.0 to +7.5 pp and the GPT-5.5 interval from $-$4.4 to +6.9 pp. Both include zero and both include premiums of several points in either direction. Under the all-graded convention the differences are $-$1.25 and +0.6 pp. Blending the screening run in as a third repeat for the native cells, the design's original intent before A3, gives $-$2.9 pp [$-$11.7, +5.9] and +1.9 pp [$-$2.7, +6.4], with the caveat that the screening run also produced the pool. The GEE logistic model the protocol's H2 calls for (pass \textasciitilde{} nativeness $\times$ vendor + workload, exchangeable within-task correlation, robust standard errors, substituting for the planned random-intercept model) estimates a nativeness $\times$ vendor interaction of 0.12 on the logit scale (robust SE 0.23, p = 0.60). The harness effect does not differ detectably between vendors. On the tasks with two graded pass/fail repeats in both cells of a contrast (78 for Opus, 77 for GPT-5.5), pass@2 is 57.7\% for both C1 and C2, and 61.0\% against 57.1\% for C3 against C4, and the fraction of tasks solved on both repeats is 42.3\% against 44.9\% (Opus) and 46.8\% against 48.1\% (GPT-5.5). Counting every missing or non-pass repeat as a failure over all 80 tasks gives pass@2 of 56.3\% for both Opus cells and 62.5\% against 58.8\% for the GPT-5.5 cells.

These intervals are the statement of what the study can and cannot say. They do not establish equivalence, and they do not give a transportable probability that a harness change moves solve rates by any particular amount on other work. The planned detectable effect of \S2.3 is a design quantity, not an observed bound.

\subsection{The workload split}

The per-family solve rates (\S4.5) show the Opus pair moving in opposite directions on contest and repository tasks. Because H3 of the plan anticipated concentration of any premium by task type, we test the split, and we label it as chosen after seeing the panel. The partition into repository (61 tasks) and contest (19 tasks) is one of several partitions the panel invites.

\begin{table}[t]
\centering\small
\caption{Paired contrasts by workload (exploratory)}
\label{tab:4-3}
\footnotesize\setlength{\tabcolsep}{4pt}
\sbox{\tblbox}{%
\begin{tabular}{llrrrrll}
\toprule
Model & Stratum & $n$ & Native & Neutral & Diff. (pp) & Task-bootstrap 95\% CI & $p$ \\
\midrule
Opus 4.8 & repository & 61 & 37.7\% & 46.7\% & -9.0 & [-17.2, -0.8] & 0.061 \\
Opus 4.8 & contest & 19 & 84.2\% & 60.5\% & +23.7 & [+2.6, +44.7] & 0.092 \\
Opus 4.8 & interaction & & & & -32.7 & [-56.7, -9.7] & 0.003 labels, 0.014 strata \\
GPT-5.5 & repository & 61 & 41.8\% & 40.2\% & +1.6 & [-5.7, +9.0] & 0.826 \\
GPT-5.5 & contest & 19 & 100.0\% & 100.0\% & +0.0 & n/a & n/a \\
GPT-5.5 & interaction & & & & +1.6 & [-5.7, +9.0] & n/a \\
\bottomrule
\end{tabular}}%
\ifdim\wd\tblbox>\linewidth\resizebox{\linewidth}{!}{\usebox{\tblbox}}\else\usebox{\tblbox}\fi

\end{table}

\begin{figure}[t]
\centering
\includegraphics[width=0.92\linewidth]{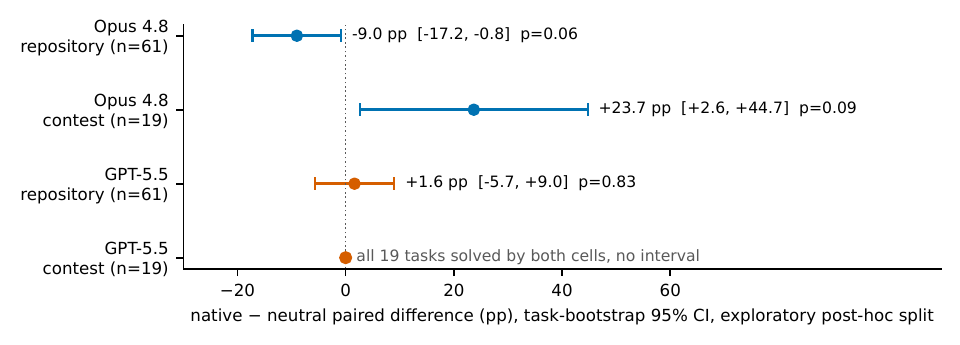}
\caption{Exploratory post-hoc split of the paired contrasts by workload (repository vs contest tasks), task-bootstrap 95\% intervals. On Opus 4.8 the sign of the harness effect reverses between workloads. On GPT-5.5 both harnesses solve every contest task.}
\label{fig:workload}
\end{figure}

On Opus 4.8 the native harness trails the neutral one on repository tasks and leads it on contest tasks. The two strata combine exactly into the overall estimate: 61/80 $\times$ ($-$9.0 pp) + 19/80 $\times$ (+23.7 pp) = $-$1.25 pp. The overall figure is therefore an average of two effects of opposite sign, weighted by this pool's workload mix. The difference of differences is $-$32.7 pp with a task-bootstrap interval that excludes zero. Two permutation tests address it, with different nulls. The primary test for H3 shuffles the workload labels across tasks while keeping each task's paired difference. Its null is that the label carries no information about the effect, which is the interaction null, and it gives p = 0.003. The stratified sign-flip test flips the sign of each task's difference within its stratum. Its null is that the harness has no effect in either stratum, and it gives p = 0.014. The per-vendor GEE agrees on the logit scale (nativeness $\times$ contest 1.62, robust SE 0.66, p = 0.014, and nativeness within repository $-$0.37, p = 0.035). These are different estimands on different scales and we report them as convergent descriptions, not as one confirmed effect. With Benjamini–Hochberg control over the four primary tests (H1 for each vendor, H2, and H3 for Opus), the interaction has q = 0.013 and nothing else approaches significance. The first version of this revision controlled a seven-test family that mixed primary and supplementary tests, under which the same interaction had q = 0.10, and we report both because the family changed after a review. On GPT-5.5 both harnesses solve every contest task, so no interaction is estimable in that stratum and the repository difference is null. Figure~\ref{fig:workload} shows the split.

We report the pattern as a post-hoc finding with two properties. It is large enough to matter for any deployment whose workload resembles one stratum more than the other, and it is not a confirmed population interaction, because the partition was one of several the family panel invited and was chosen after seeing it. \S6 discusses what follows for practitioners.

\subsection{Correct output and autonomous completion}

The August manuscript treated the wall-clock ceiling as a graded failure. The archive shows otherwise. The driver cancels the agent at 1,200 s and the oracle grades the patch present at cancellation. Of 81 graded runs that hit the ceiling, \textbf{22 passed}: C2 11 of 32, C4 4 of 7, C5 2 of 12, C6 5 of 29, and C1 0 of 1. Correctness and autonomous completion are therefore distinct endpoints in our data. The neutral harness reached the ceiling far more often (C2: 32 of 160 final attempts against C1: 1. C4: 7 against C3: 0), which is a latency and cost fact rather than a capability one. Median wall-clock for completed runs was 392 s (C1), 596 s (C2), 280 s (C3), 420 s (C4), 650 s (C5) and 943 s (C6). The neutral harness also issued about twice as many tool invocations per session as its native counterpart (medians C1 44.5 against C2 90.5, and C3 51.5 against C4 75). This is a descriptive count of \texttt{tool\_\allowbreak{}call\_\allowbreak{}started} events and not a validated count of model calls. Whether a correct patch existed earlier in a ceilinged trajectory, and whether stopping could be optimized, are questions for trajectory-level measurement that this archive does not support.

\begin{table}[t]
\centering\small
\caption{Outcome flow, planned key $\rightarrow$ verdict (final attempts)}
\label{tab:4-4}
\footnotesize\setlength{\tabcolsep}{4pt}
\sbox{\tblbox}{%
\begin{tabular}{lrrrrrll}
\toprule
Cell & completed & ceiling & agent\_error & infra-terminal & Graded & pass / fail / timeout / apply\_error & Ceiling runs passed \\
\midrule
C1 & 150 & 1 & 8 & 1 & 159 & 78 / 81 / 0 / 0 & 0 of 1 \\
C2 & 126 & 32 & 2 & 0 & 160 & 80 / 79 / 0 / 1 & 11 of 32 \\
C3 & 160 & 0 & 0 & 0 & 160 & 87 / 71 / 2 / 0 & n/a \\
C4 & 152 & 7 & 1 & 0 & 160 & 86 / 73 / 1 / 0 & 4 of 7 \\
C5 & 67 & 12 & 1 & 0 & 80 & 35 / 43 / 2 / 0 & 2 of 12 \\
C6 & 14 & 29 & 30 & 7 & 73 & 14 / 57 / 2 / 0 & 5 of 29 \\
\bottomrule
\end{tabular}}%
\ifdim\wd\tblbox>\linewidth\resizebox{\linewidth}{!}{\usebox{\tblbox}}\else\usebox{\tblbox}\fi

\end{table}

\subsection{Side cells, screening agreement, and per-family structure}

The four frontier cells lie between 48.8\% and 55.6\%. C5 (gemini-3.5-flash) lands at 44.9\%. C6 (deepseek-v3.2 via DeepInfra) reaches 19.7\%, and its shortfall is operational as much as it is capability: 30 of its 80 final attempts ended in agent errors and 29 at the ceiling, leaving 14 completed runs. The screening pass measured agreement between the two vendor-native bundles (model and harness together, not the harness alone): C1 and C3 agreed on 91\% of the 256 tasks (132 both-pass, 100 both-fail, 24 discordant), with screening rates of 54.3\% (C1) and 58.2\% (C3). The per-family panel of the frozen pool (Appendix C of the preprint) shows the Opus pair at 39\% against 43\% on AccessCtl, 27\% against 42\% on IdentityApp, 46\% against 62\% on FieldSvc and 84\% against 61\% on the contest tasks, which is the pattern \S4.3 examined.

\section{Cost Results, Telemetry Correction, and Calibration}

\subsection{A usage-semantics defect in the telemetry path}

Cost was a research question (RQ2, H4), and the plan's cost basis was to store raw tokens per run and price them in analysis from a versioned table, with SDK-reported cost as a cross-check. The pipeline that implemented it has three stages (Figure~\ref{fig:arch}). The runner inside the VM emits per-turn usage on its \texttt{result\_\allowbreak{}done} message. The host normalizer (\texttt{app/\allowbreak{}sessions/\allowbreak{}telemetry.py}) folds it into \texttt{tokens\_\allowbreak{}in}, \texttt{tokens\_\allowbreak{}out} and cache counters accumulated on the session row. The ledger copies the session totals and prices them (\texttt{eval/\allowbreak{}orchestrator/\allowbreak{}pricing.py}). The normalizer contains one rule that is correct for exactly one of the three harnesses: if the usage record carries cache-read or cache-write keys, add both to \texttt{tokens\_\allowbreak{}in}.

Anthropic's native API reports \texttt{input\_\allowbreak{}tokens} exclusive of cache reads and writes, so for claude-agent-sdk the rule produces the correct inclusive total. The openai-codex SDK reports \texttt{input\_\allowbreak{}tokens} inclusive of cached input under a differently named key that the rule does not match, so nothing is added, which is also correct. LangChain's \texttt{usage\_\allowbreak{}metadata}, which deepagents surfaces, reports \texttt{input\_\allowbreak{}tokens} inclusive of cache, and our deepagents runner copied its cache details into the Anthropic-style keys. The rule then added the cache tokens a second time. For a trajectory with 1,000 inclusive input tokens of which 900 were cache reads and 50 cache writes, the ledger records 1,950 input tokens, and a cache-read share computed from it reads 46\% instead of 90\%. Table~\ref{tab:5-2} reconciles every main-run ledger row against the raw per-turn events in the archive.

\begin{table}[t]
\centering\small
\caption{Source-to-ledger reconciliation, main run}
\label{tab:5-2}
\footnotesize\setlength{\tabcolsep}{4pt}
\sbox{\tblbox}{%
\begin{tabular}{lrrrrrr}
\toprule
Cell & Rows with usage / rows & Ledger $=$ source total & Ledger $=$ total $+$ R $+$ W & Other & Share as reported & Share, source semantics \\
\midrule
C1 & 152 / 185 & 152 & (same) & 0 & 97.0\% & 97.0\% \\
C2 & 147 / 172 & 0 & 138 & 9 & 48.9\% & 97.7\% \\
C3 & 160 / 160 & 160 & n/a & 0 & 93.9\% & 93.9\% \\
C4 & 154 / 160 & 0 & 153 & 1 & 48.0\% & 92.4\% \\
C5 & 74 / 83 & 0 & 74 & 0 & 46.6\% & 87.1\% \\
C6 & 16 / 131 & 0 & 16 & 0 & 48.1\% & 92.6\% \\
\bottomrule
\end{tabular}}%
\ifdim\wd\tblbox>\linewidth\resizebox{\linewidth}{!}{\usebox{\tblbox}}\else\usebox{\tblbox}\fi

\end{table}

The reconciliation is exact for 693 of the 703 rows that have raw usage. The ten remaining rows (nine C2, one C4) belong to sessions with three or more \texttt{result\_\allowbreak{}done} events and sit within $-$2.6\% to +0.2\% of the double-added total, consistent with the host accumulator losing part of one turn's increment under concurrent re-entry. Rows without raw usage are runs that ended before any \texttt{result\_\allowbreak{}done}. They carry no tokens in either source, which is why ledger-side spend estimates are lower bounds (\S5.3). C6 has raw usage for only 16 of 131 rows, so its cost is not estimable.

Under source semantics every cell caches at 87–98\% (Table~\ref{tab:5-2}, and the figure in Appendix D of the preprint). The neutral harness is in the same regime as the native harnesses, and the 48\% figure of the August manuscript was the arithmetic signature of the double count. A second, independent pricing error compounded it: the ledger formula subtracted cache reads from \texttt{tokens\_\allowbreak{}in} but not cache writes, so writes were charged at the input rate and again at the cache-write rate on every harness. Both are fixed in the platform with regression tests. The historical analysis below re-prices from the raw events rather than patching the ledger.

\subsection{Cost per solved task from observed usage}

We re-price every run from the raw per-turn usage with an explicit formula. Uncached input (inclusive total minus reads minus writes) is charged at the input rate, cache reads at the cache-read rate, cache writes at the cache-write rate and output at the output rate, all from the frozen pricing table (v2, 2026-07-13). Table~\ref{tab:5-3} gives the corrected estimate per cell with its coverage, and for C1 the cost the harness's own SDK reports. The original ledger totals are in Appendix D of the preprint. Figure~\ref{fig:cost} plots the per-cell cost per solved task.

\begin{table}[t]
\centering\small
\caption{Corrected cost per cell, main run (USD, list prices)}
\label{tab:5-3}
\footnotesize\setlength{\tabcolsep}{4pt}
\sbox{\tblbox}{%
\begin{tabular}{llrrrrl}
\toprule
Cell & Harness / model & Rows with usage / rows & Corrected estimate (\$) & SDK-reported (\$) & Passes & Corrected \$ per solve \\
\midrule
C1 & claude-sdk / Opus 4.8 & 152 / 185 & 242 & 295 & 78 & 3.10 (3.79 by SDK) \\
C2 & deepagents / Opus 4.8 & 147 / 172 & 406 & n/a & 80 & 5.07 \\
C3 & codex-sdk / GPT-5.5 & 160 / 160 & 251 & n/a & 87 & 2.88 \\
C4 & deepagents / GPT-5.5 & 154 / 160 & 292 & n/a & 86 & 3.39 \\
C5 & deepagents / Gemini 3.5 Flash & 74 / 83 & 164 & n/a & 35 & 4.67 \\
C6 & deepagents / DeepSeek V3.2 & 16 / 131 & not estimable & n/a & 14 & not estimable \\
\bottomrule
\end{tabular}}%
\ifdim\wd\tblbox>\linewidth\resizebox{\linewidth}{!}{\usebox{\tblbox}}\else\usebox{\tblbox}\fi

\end{table}

\begin{figure}[t]
\centering
\includegraphics[width=\linewidth]{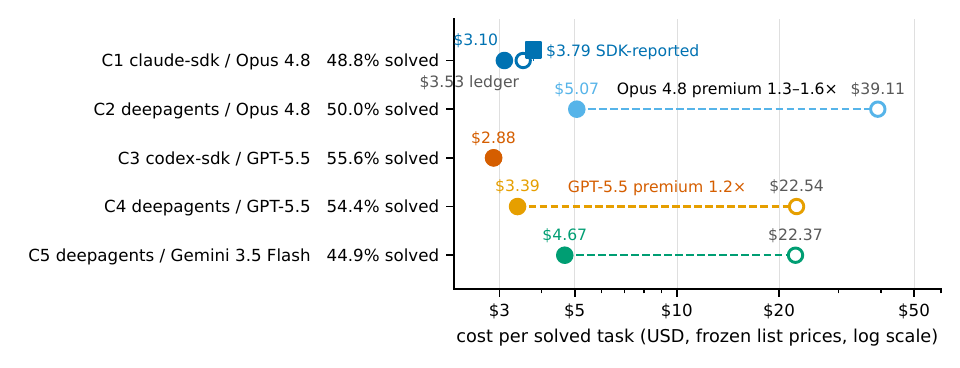}
\caption{Cost per solved task by cell (USD at frozen list prices, log scale). Rows carry the cell's solve rate. Hollow markers show the original ledger, filled markers the estimate re-priced from raw per-turn usage, and the square C1 at the cost the native SDK itself reports. C6 is omitted (usage recorded for 16 of 131 rows).}
\label{fig:cost}
\end{figure}

The cost ratio of a contrast carries three separate uncertainties, and we quantify each. The first is the pricing basis. Anthropic's SDK reports C1's cost at \$295, which is 22\% above our list-price estimate of \$242. Holding input, read and output prices at list, the SDK figures imply an effective cache-write rate of about \$14 per million tokens against the table's \$6.25, or usage that the \texttt{result\_\allowbreak{}done} record does not surface. At list prices for both cells the Opus ratio is 1.63. With C1 at its SDK-reported cost it is 1.34, and with both cells at the 1-hour cache-write rate it is 1.5–1.6. The GPT-5.5 ratio is 1.18 and has no cache-write term. The second uncertainty is task sampling. Table~\ref{tab:5-6} gives task-bootstrap intervals for the observed-cost ratio in which all attempts of a task are resampled together: [1.29, 2.09] for Opus at list prices, [1.06, 1.71] with C1 at SDK cost, and [1.05, 1.33] for GPT-5.5. The third uncertainty is usage that was never recorded, which \S5.3 quantifies and which is material on the Anthropic account. We therefore report the Opus ratio as \textbf{1.3–1.6 on observed usage} and the GPT-5.5 ratio as \textbf{1.2}, and we do not present either as a bound on the billed ratio. The 2.6 and 1.8 ratios of the August manuscript were obtained by dividing dashboard balance deltas by ledger totals inflated by the defect, and they are withdrawn.

Table~\ref{tab:5-4} decomposes each cell's corrected cost into its four components. The native Opus cell has almost no uncached input, because claude-agent-sdk writes the whole prefix to cache on the first call and reads it back on every later call. Its cost is 58\% cache reads, 17\% cache writes and 25\% output, and the neutral Opus cell has the same structure at a larger volume (65\% reads, 14\% writes, 21\% output). The Opus gap of \$164 is \$124 in cache reads, \$17 in cache writes and \$23 in output. On GPT-5.5 both cells spend about a quarter on uncached input, half on cache reads and a fifth on output, and the gap of \$41 is \$17 in uncached input, \$19 in reads and \$5 in output. The median per-run inclusive input (1.04 Mtok on C1 against 2.52 on C2, and 1.34 on C3 against 1.32 on C4) is consistent with this but does not by itself decompose a total, as the GPT-5.5 pair shows.

\begin{table}[t]
\centering\small
\caption{Corrected cost by component, main run (USD at list prices, share of the cell total)}
\label{tab:5-4}
\footnotesize\setlength{\tabcolsep}{4pt}
\sbox{\tblbox}{%
\begin{tabular}{lrrrrrr}
\toprule
Cell & Uncached input & Cache reads & Cache writes & Output & Total (\$) & Rows with usage / rows \\
\midrule
C1 & 0 (0\%) & 140 (58\%) & 41 (17\%) & 61 (25\%) & 242 & 152 / 185 \\
C2 & 0 (0\%) & 263 (65\%) & 58 (14\%) & 84 (21\%) & 406 & 147 / 172 \\
C3 & 67 (27\%) & 132 (53\%) & 0 (0\%) & 52 (21\%) & 251 & 160 / 160 \\
C4 & 84 (29\%) & 151 (52\%) & 0 (0\%) & 56 (19\%) & 292 & 154 / 160 \\
C5 & 80 (49\%) & 62 (38\%) & 0 (0\%) & 22 (13\%) & 164 & 74 / 83 \\
C6 & \multicolumn{4}{l}{not estimable (16 of 131 rows carry usage)} & 1.6 & 16 / 131 \\
\bottomrule
\end{tabular}}%
\ifdim\wd\tblbox>\linewidth\resizebox{\linewidth}{!}{\usebox{\tblbox}}\else\usebox{\tblbox}\fi

\end{table}

\subsection{What billed spend does and does not confirm}

The PI recorded exact account balances at the stop-loss halt (2026-08-06) and after the finish leg, with no top-ups in between. These are observations, not ledger-derived, and they are the only external check on the corrected estimator. Table~\ref{tab:5-5} places them beside the ledger-side estimates for two candidate windows: rows whose \texttt{finished\_\allowbreak{}at} precedes the halt record, and all main-run rows. The balance readings were recorded to the day.

\begin{table}[t]
\centering\small
\caption{Account-level reconciliation (USD)}
\label{tab:5-5}
\footnotesize\setlength{\tabcolsep}{4pt}
\sbox{\tblbox}{%
\begin{tabular}{llrrrrr}
\toprule
Account (cells) & Window & Balance delta & Original ledger & Corrected estimate & Deviation & Rows without usage \\
\midrule
OpenAI (C3 $+$ C4) & halt window & 347.42 & 1,491 & 333.38 & -4.0\% & 6 \\
OpenAI (C3 $+$ C4) & all main rows & 553.62 & 2,189 & 542.72 & -2.0\% & 6 \\
Anthropic (C1 $+$ C2) & halt window & 522.51 & 1,709 & 380.09 & -27\% & 28 \\
Anthropic (C1 $+$ C2) & all main rows & 989.44 & 3,404 & 701.02 & -29\% & 58 \\
\bottomrule
\end{tabular}}%
\ifdim\wd\tblbox>\linewidth\resizebox{\linewidth}{!}{\usebox{\tblbox}}\else\usebox{\tblbox}\fi

\end{table}

The raw events identify what each recorded turn consumed. On the OpenAI account, where only 6 rows lack a usage record, the corrected estimator reproduces the billed spend within 2–4\% in both windows. That is an account-level check and not a per-cell billing validation, but it shows the corrected accounting is right in kind and magnitude where coverage is complete. On the Anthropic account a residual of 25–29\% remains in both windows and does not close under the 1-hour cache-write assumption ($-$24\% and $-$26\%). The residual has the sign and roughly the size of the spend the ledger cannot see: 58 Anthropic rows (33 C1, 25 C2) ended without any \texttt{result\_\allowbreak{}done}, against 6 such rows on OpenAI. Table~\ref{tab:5-7} shows what the C2/C1 per-solve ratio would be under four allocations of the full-window residual of \$288: none, all to C1, all to C2, and in proportion to each cell's rows without usage. The ratio moves from 1.34 with no allocation to 0.68 if the residual all belongs to C1 and 2.29 if it all belongs to C2, and to 1.13 under the row-proportional rule. These are illustrations, not estimates, and they assume the balance windows and the ledger windows coincide to the day. They show that the billed ordering of the two Opus cells is unresolved by the evidence we have. Across all phases, the corrected ledger estimate of the study's provider spend is about \$2.3K (main about \$1.4K, screening about \$0.8K, dress, probe and smokes about \$0.1K), against \$2.4–2.6K by balance deltas plus a postpaid Google invoice estimated at about \$250.

\begin{table}[t]
\centering\small
\caption{Observed-cost ratio per solved task with task-bootstrap intervals}
\label{tab:5-6}
\footnotesize\setlength{\tabcolsep}{4pt}
\begin{tabular}{llrl}
\toprule
Ratio & Pricing basis & Observed usage & Task-bootstrap 95\% CI \\
\midrule
C2 / C1 & both at list prices & 1.63 & [1.29, 2.09] \\
C2 / C1 & C1 at SDK-reported cost & 1.34 & [1.06, 1.71] \\
C4 / C3 & both at list prices & 1.18 & [1.05, 1.33] \\
\bottomrule
\end{tabular}

\end{table}

\begin{table}[t]
\centering\small
\caption{Allocation of the unrecorded Anthropic spend (USD) and the resulting C2/C1 ratio}
\label{tab:5-7}
\footnotesize\setlength{\tabcolsep}{4pt}
\sbox{\tblbox}{%
\begin{tabular}{lrrrrrl}
\toprule
Window & Balance $\Delta$ & C1 obs. (SDK) & C2 obs. & Residual & Rows w/o usage & Ratio none / to C1 / to C2 / by rows \\
\midrule
halt window & 522.51 & 175.63 & 204.46 & 142.42 & 14 / 14 & 1.14 / 0.63 / 1.93 / 1.09 \\
all main rows & 989.44 & 295.28 & 405.73 & 288.42 & 33 / 25 & 1.34 / 0.68 / 2.29 / 1.12 \\
\bottomrule
\end{tabular}}%
\ifdim\wd\tblbox>\linewidth\resizebox{\linewidth}{!}{\usebox{\tblbox}}\else\usebox{\tblbox}\fi

\end{table}

\subsection{The truncation probe}

After the finish leg we ran a probe of one configuration hypothesis: deepagents' default tool-output truncation (100,000 bytes) against a native-like 30,000-byte cap, on 12 repository tasks under C2 and C4, one run each. The August manuscript reported the probe as having falsified the truncation mechanism because cost, tokens and the 48\% cache share were invariant. Four facts change that reading. The 48\% share was the defect, so its invariance was uninformative. The probe differed from the main run in more than the knob: concurrency 8 rather than 12, one repeat rather than two, and a different day. The bootstrap intervals on the per-run cost difference (Table~\ref{tab:5-8}) span roughly $-$19\% to +6\% of the main-run mean for C2 and $-$20\% to +53\% for C4, so the probe cannot distinguish a moderate saving from a moderate increase. And whether the 30 KB cap ever bound cannot be established, because both the canonical events and the wire traces retain tool results only as 2,000-character previews. We inspected 2,868 and 2,457 main-run tool results and 1,333 and 1,331 probe results, and none is longer than the preview, so the archive holds no record of raw tool-output sizes.

\begin{table}[t]
\centering\small
\caption{Truncation probe at the 12 matched tasks (corrected, USD per run)}
\label{tab:5-8}
\footnotesize\setlength{\tabcolsep}{4pt}
\sbox{\tblbox}{%
\begin{tabular}{lrrlllr}
\toprule
Cell & Main \$/run & Probe \$/run & Probe $-$ main (95\% CI) & Input (Mtok) & Cache share & Solve \\
\midrule
C2 / C2p & 3.68 & 3.43 & -0.25 [-0.70, +0.22] & 3.47 $\rightarrow$ 3.00 & 98.0\% $\rightarrow$ 97.9\% & 54\% $\rightarrow$ 58\% \\
C4 / C4p & 2.13 & 2.43 & +0.30 [-0.43, +1.13] & 1.37 $\rightarrow$ 1.61 & 93.7\% $\rightarrow$ 95.2\% & 58\% $\rightarrow$ 58\% \\
\bottomrule
\end{tabular}}%
\ifdim\wd\tblbox>\linewidth\resizebox{\linewidth}{!}{\usebox{\tblbox}}\else\usebox{\tblbox}\fi

\end{table}

We report the probe as an inconclusive pilot. The trajectory-level explanation the August manuscript advanced in its place, more model calls over a growing context, is consistent with the volume and tool-invocation counts but was not tested by any intervention, and we do not claim it as a mechanism. The methodological takeaway is concrete: a cap experiment needs demonstrated exposure to the cap, which means retaining raw tool-output lengths and truncation events in the trace.

\section{Discussion}

\textbf{Workload composition changes the aggregate comparison.} The Opus contrast is $-$1.25 pp overall, and \S4.3 showed that this number is 61/80 of a $-$9.0 pp repository effect plus 19/80 of a +23.7 pp contest effect. A team whose work resembles the repository stratum would read this as weak evidence that the neutral harness solves more, and a team with short self-contained problems would read it the other way. Neither reading is confirmed. The partition was chosen after seeing the panel, the strata intervals are wide, and a permutation p of 0.003 on a post-hoc partition is evidence for a designed test rather than a result of one. What the pattern does establish is that a single aggregate solve rate is not the quantity a deployment decision needs. The decision needs the harness effect under the user's own workload mix, with its uncertainty. A designed replication would pre-specify the strata, power the contest stratum (19 tasks here) and use three repeats.

\textbf{Correct output and autonomous completion are different endpoints.} The oracle grades patches, and 22 of 81 runs cancelled at the ceiling had produced a passing patch. The neutral harness hit the ceiling far more often and issued about twice the tool invocations, at similar solve rates. A benchmark that scores only the patch will call these harnesses equivalent, and a user waiting for the agent to finish will not. The two endpoints should be reported together: solve rate, the distribution of terminal states, and time to completion. Whether a correct patch was available earlier in those trajectories, and whether stopping can be optimized safely, are questions for trajectory-level measurement.

\textbf{Measured token cost and billed operational cost need different evidence.} The raw per-turn events identify what every recorded turn consumed, and re-pricing them at frozen list prices gives the observed-usage ratios of \S5.2. Billed spend is a different quantity. It includes turns that never produced a usage record and it depends on the provider's actual rates, including cache-write terms our list-price table did not resolve. On the OpenAI account the two quantities agree within 2–4\%, because almost every run left a record. On the Anthropic account 58 runs left none, and allocating the 25–29\% residual to either cell moves the ratio between 0.7 and 2.3, which is enough to reverse the ordering of the two Opus cells. The lesson for anyone pricing agent runs from harness telemetry has two parts. Pin the usage semantics of every SDK before applying a price, because inclusive and exclusive input counts differ by exactly the cache tokens. Then reconcile the total against money that left the account and treat any residual as unallocated spend.

\textbf{A cap experiment needs demonstrated exposure to the cap.} The truncation probe could not show that its treatment ever applied, because neither the canonical events nor the wire traces kept raw tool-output lengths. An evaluation that varies a truncation, summarization or eviction setting should log, per tool call, the raw output length and whether the policy fired. Without that record an invariant outcome is uninterpretable.

\textbf{Relation to concurrent and subsequent work.} Three 2026 studies address harness effects with different designs (\S8). Harness-Bench treats its results as diagnostics of harness configurations and reports substantial variation in completion as well as in process measures. The Scaffold Effect compares three open-source harnesses and finds paired within-model pass-rate differences of 0–8 pp with intervals that include zero, alongside a 40$\times$ spread in tokens per solved task. Same Model, Different Harness studies a conditional intervention, shortening older tool results as the context fills, and finds a large effect under a tight context window and almost none under a wide one. None of these is the vendor-native against neutral comparison we ran, and we do not read them as votes for an average null. They motivate what our data also suggests: the harness effect is conditional on workload, context regime and the endpoint measured, so studies should fix those conditions and name their comparators. In August 2026, after our study, HarnessRouter published a single-task benchmark comparing its Hermes harness with vendor harnesses on the same models and reported Hermes as 1.5–2.1 times cheaper \cite{harnessrouter2026}. Its costs are expressed in application credits rather than in model-usage dollars, its design is one synthetic task with five runs per configuration and no contamination control, and our observed-usage estimate points the other way. The two are not comparable, and together they show that the sign of the harness cost effect is not settled.

\textbf{What a practitioner can take from this study.} On an 80-task private pool, two same-model contrasts did not resolve an average solve-rate advantage for either harness, with intervals of roughly 11 and 17 pp width. The neutral harness took longer, reached the ceiling more often and, on observed usage at list prices, cost 1.2–1.6 times as much per solved task, with the Anthropic ordering unresolved by billing. The choice between a vendor harness and a portable one should therefore be made on the user's workload mix, on the completion behaviour the user needs, and on billed cost under the user's own pricing terms, not on an aggregate solve rate.

\section{Threats to Validity and Limitations}

\textbf{Task provenance and selection.} All 179 Track A tasks were mined from four codebases within one organization, and the frozen pool's family mix (contest 19, AccessCtl 36, IdentityApp 13, FieldSvc 12) reflects our mining yield rather than the population of software work. The pool oversamples screening-discordant tasks (30\% of the pool against 9.4\% of the suite), so every main-run average is conditional on that selection (\S2.3). The white-box coupling of roughly 30 Track A tasks cannot be assessed because the per-task labels were not retained. Shared internal symbols could favour one harness's behaviour over another's, and we cannot rule that out.

\textbf{Configuration differences beyond the harness.} The codex cell ran in 4 GiB VMs and the other cells in 8 GiB (Appendix B). Inference budgets and tool surfaces also differ, because identical tools cannot be imposed across independently engineered products. The contrasts compare deployed configurations of each harness on the same model.

\textbf{Power and inference.} Cost constraints reduced C2/C4 to two repeats and the pool to 80 tasks (A3, A4). The task-bootstrap intervals span 11–17 pp. The planned detectable effect (8–10 pp, about 12 pp after A3) is a design quantity and is not an observed bound. The run-pair McNemar test is not task-clustered and is supplementary, and the cluster-adjusted version agrees with the bootstrap. The two interaction tests have different nulls and the GEE interaction a different scale. The GEE model substitutes for the planned random-intercept model, and the GPT-5.5 contest stratum is fully separated, so its logistic coefficients are not estimable. Side cells run at k = 1.

\textbf{Post-hoc analyses.} The workload split (\S4.3), the outcome-flow reanalysis (\S4.4) and the probe re-reporting (\S5.4) were conducted in September 2026 on the archived data. Only the H1 to H4 tests were specified in advance. Every post-hoc result is labelled. False-discovery control is applied over one primary test per hypothesis, and the family was changed once after a review, which \S4.3 discloses with both q-values. The number of partitions considered before choosing repository against contest is not recoverable.

\textbf{Cost measurement.} The corrected estimator is validated against billed spend on one provider (OpenAI, within 2–4\%) and leaves a 25–29\% residual on the other (Anthropic), plausibly but not verifiably attributable to runs that ended without a usage record. The Opus cost ratios are observed-usage estimates with task-bootstrap intervals under stated pricing assumptions, and the billed ordering is unresolved (Table~\ref{tab:5-7}). Cache-write pricing for Anthropic is uncertain by a factor of about 2 (5-minute against 1-hour rates against the rate implied by the SDK). C6 cost is not estimable. All prices are list prices frozen on 2026-07-13, and C6 is priced at DeepSeek list rather than DeepInfra.

\textbf{Termination and ceilings.} 81 graded runs hit the 1,200 s ceiling and 22 of them passed. The ceiling censors time rather than correctness, and a longer ceiling could change both solve rates and costs for the neutral harness, which hit it far more often. The interim results at the halt differed from the final ones by several points in both directions (\S4.1).

\textbf{Execution.} The main run halted at its stop-loss after 502 rows and was completed in a finish leg two days later under amended caps (A6), with a deploy freeze, pinned session configuration and the drift gate re-passing before resume. The sandbox image was not pinned by OCI digest in the ledger (\texttt{image\_\allowbreak{}digest: unpinned} on every row). The archive records a build fingerprint of the image source on 889 of 891 sessions and the library versions on every session. Thirteen C1 runner deaths on 8 GB VMs remain unexplained. All were scrubbed and retried and none was graded.

\textbf{Reproducibility after teardown.} The per-task mirrors were deleted after the study. Track A tasks are reconstructable from the private source repositories at the recorded commits (24 of 24 sampled commits resolved in August 2026), and Track B hidden tests are held privately. The checksummed archive of ledgers, verdicts, per-run traces and per-turn events supports every number in this paper but not a re-run of the agents against the tasks.

\section{Related Work}

\textbf{Execution-based coding benchmarks.} SWE-bench\cite{jimenez2024swebench} established repository-level issue resolution graded by hidden tests. We inherit its FAIL\_TO\_PASS and PASS\_TO\_PASS discipline for the 179 repository tasks and enforce gold-passes and base-fails polarity as an automated gate over all 256 environments. SWE-bench Verified\cite{openai2024verified} showed that task quality is an empirical claim requiring validation, which motivated our prompt-finalization protocol. LiveCodeBench\cite{jain2024livecodebench} established the rolling post-cutoff pattern our 77 contest tasks adopt.

\textbf{Contamination.} Longitudinal studies show performance discontinuities at the training cutoff\cite{roberts2024cutoff}, and SWE-bench-specific analyses report solution leakage\cite{aleithan2024swebenchplus}. The standard mitigations are private suites, post-cutoff harvesting and canary strings\cite{srivastava2023bigbench}. We deploy all three behind a frozen, machine-readable cutoff registry and add a runtime drift gate, because providers swap dated snapshots behind stable aliases. \S2.2 bounds the resulting claim.

\textbf{Harness comparisons.} SWE-agent\cite{yang2024sweagent} showed that scaffold design is a first-order variable. OpenHands\cite{wang2024openhands} provides a platform for comparing agent designs, terminal-bench\cite{tbench2025} publishes leaderboard rows as (harness, model) pairs, and AgentBench\cite{liu2024agentbench} varies models under a fixed scaffold. Three studies concurrent with ours address the harness question directly, each with a different comparator. Harness-Bench\cite{yao2026harnessbench} (May 2026) evaluates harness configurations across model backends on 106 sandboxed tasks and 5,194 trajectories. It presents its results as diagnostics of configurations and reports substantial variation in completion, process quality, efficiency and failure behaviour, and it argues that capability should be attributed to model and harness pairings. Vats and Golev's Scaffold Effect\cite{vats2026scaffold} (June 2026) compares three open-source harnesses on 50 tasks and finds up to a 40$\times$ spread in tokens per solved task while paired within-model pass-rate differences stay at 0–8 pp with bootstrap intervals that include zero except for the largest gap. Lewis's Same Model, Different Harness\cite{lewis2026sameharness} (August 2026) evaluates one harness intervention, shortening older tool results as the context fills, and finds that it roughly doubles fail-to-pass fractions under a tight context window and does almost nothing under a wide one. Our study differs in comparator and setting: the vendors' own production harnesses against one neutral harness, on the same model, on a private contamination-controlled suite, with cost reconciled against billed spend. A subsequent industry benchmark from HarnessRouter (August 2026)\cite{harnessrouter2026} compares its Hermes harness with vendor harnesses on one synthetic task and reports Hermes as cheaper in application credits. \S6 explains why that result and ours are not comparable.

\textbf{Inference economics and usage accounting.} Kapoor et al.\cite{kapoor2024agents} argue that agent leaderboards should report cost alongside accuracy. Where the cost numbers come from has had little scrutiny. Prompt caching\cite{vendors2026caching} makes effective input price a function of trajectory shape, and, as \S5.1 shows, it also makes usage records provider- and SDK-specific: Anthropic's API reports input exclusive of cache, and OpenAI's and LangChain's aggregates report it inclusive\cite{vendors2026usage}. We are not aware of prior work documenting that a multi-SDK telemetry path can double-count cache tokens for some harnesses and not others, or of a published agent-cost comparison reconciled against provider balances.

\section{Conclusion}

We asked whether a vendor's own agent harness solves more tasks than a neutral multi-provider harness on the same model, and we measured it with paired contrasts on a private, contamination-controlled pool of 80 tasks. Neither contrast resolved an average advantage. The Opus 4.8 difference is $-$1.25 pp with a task-bootstrap interval of [$-$10.0, +7.5], and the GPT-5.5 difference is +1.25 pp with an interval of [$-$4.4, +6.9], conditional on this pool. On Opus the average combines a repository stratum favouring the neutral harness with a contest stratum favouring the native one, a post-hoc pattern with a permutation p of 0.003 that a designed replication should test. The two harnesses differ clearly in completion behaviour and in observed usage: the neutral harness reached the ceiling more often, made about twice the tool calls, and on observed usage at list prices cost 1.2–1.6 times as much per solved task, with the billed ordering on Anthropic unresolved. The methodological result we most want readers to keep is the reconciliation discipline that this study lacked and then acquired: pin every SDK's usage semantics before pricing a token, compare raw per-turn events with the ledger, and compare the ledger with billed spend.

\section*{Data Availability}

The archived study, comprising all ledgers and per-run traces for every phase, the grading verdicts, the frozen plans and pool lock, and a CSV export of the canonical per-turn event history of every run, is retained as a sha256-manifested archive by the authors. The replication package (\ARTIFACTURL{}) contains the evaluation infrastructure (orchestrator, grading oracle, drift gate, pricing table and registry format), the reanalysis script that produces every number and figure in \S4 and \S5 from the archive, the derived per-run aggregates it emits (pseudonymised task ids, and no prompts, patches, event text or credentials), the claim-to-evidence and protocol-to-execution tables, and the revision record listing every claim of the August 2026 manuscript and its status. The private OSF project holding the June 2026 design deposit (\OSFURL{}) contains timestamped files but no formal registration.

The 256 tasks, gold patches, hidden tests and canary strings are not released. Publishing them would contaminate the post-cutoff suite, and the repository tasks derive from proprietary codebases. The per-task run mirrors were deleted after the study. Track A tasks are reconstructable from the private source repositories at the recorded commits, and Track B hidden tests are held privately by the authors. The package documents the mining, validation, registry-freeze and phase protocol in enough detail to reproduce the design on another private codebase. We will make the suite available to an artifact-evaluation committee under a non-disclosure arrangement on request.

\appendix

\section{Execution chronology and incidents (condensed)}

Dates are 2026 and times UTC. Each incident became a regression guard before the next scored phase, and no incident retroactively altered graded data.

\begin{table}[t]
\centering\footnotesize\setlength{\tabcolsep}{3.5pt}
\begin{tabularx}{\linewidth}{>{\hsize=0.26\hsize\raggedright\arraybackslash}X>{\hsize=1.74\hsize\raggedright\arraybackslash}X}
\toprule
Date & Event \\
\midrule
06-11 & Program plan (hypotheses H1–H4, cells, statistical plan) committed to the repository \\
06-24 & Cutoff registry frozen. Eligibility 2026-03-02 = binding cutoff (Opus 4.8, 2026-01-31) + 30 days \\
06-25$\rightarrow$28 & Design documents deposited to a private OSF project, including \texttt{PREREGISTRATION.md}. No registration \\
08-04 & Dress rehearsal 1 aborted: a \texttt{decimal.Decimal} from Postgres made every ledger append raise and 72 paid sessions ran with the stop-loss blind. Fix: unrecordable append halts the run \\
08-04$\rightarrow$05 & Dress rehearsal 2 overshot a €100 stop-loss to €275. Fix: predictive dispatch brake. Amendments A1–A5 \\
08-05 & Screening, C1 + C3 $\times$ 256 tasks. OpenAI credit exhaustion produced 72 consecutive C3 agent errors. Fix: scrub step. 512/512 graded, pool frozen 19:16 UTC \\
08-06 & Main run (800 keys, concurrency 12). Halted by stop-loss at 502 rows, €3,303.34 accounted, 17:50 UTC. A6 \\
08-07 & Finish leg after OpenRouter 429s blocked resume twice (drift probe pinned to DeepInfra, A7). 792/800 keys graded, 891 ledger rows. Grading disk exhaustion recovered by idempotent regrade \\
08-08 & Truncation probe, 12 tasks $\times$ \{C2p, C4p\}, 30 KB cap, concurrency 8 \\
08-10 & Teardown. Mirrors deleted, database exported, archive manifested \\
09-08 & Post-study review finds the telemetry defect. Reanalysis from the archive. This revision– \\
\bottomrule
\end{tabularx}
\end{table}

\section{Configuration summary}

YAML plans are validated by the orchestrator's closed schema, with a per-cell \texttt{sandbox} override and a \texttt{Run\allowbreak{}Spec.config\_\allowbreak{}hash} recording the effective configuration on every ledger row. All scored phases used ceiling 1,200 s, concurrency 12 (probe: 8) and \texttt{shuffle\_\allowbreak{}seed: 7}. Harness versions are those of Table~\ref{tab:2-1}, recorded on every session row. The sandbox image's OCI digest was not recorded (\texttt{image\_\allowbreak{}digest: unpinned} on all 891 ledger rows), and the session rows carry a build fingerprint of the image source (\texttt{76a4c8c5\ldots{}}) for 889 of 891 sessions. In-container verification on the probe image additionally recorded langchain 1.3.14, langchain-anthropic 1.5.3 and langchain-openai 1.4.1.

\begin{table}[t]
\centering\footnotesize\setlength{\tabcolsep}{3.5pt}
\begin{tabularx}{\linewidth}{>{\hsize=0.41\hsize\raggedright\arraybackslash}X>{\hsize=1.01\hsize\raggedright\arraybackslash}X>{\hsize=0.41\hsize\raggedright\arraybackslash}X>{\hsize=0.41\hsize\raggedright\arraybackslash}X>{\hsize=2.75\hsize\raggedright\arraybackslash}X}
\toprule
Cell & Model & Repeats & VM RAM & Key env \\
\midrule
C1 & claude-opus-4-8 & 2 & 8192 MiB & \texttt{SDK\_\allowbreak{}MAX\_\allowbreak{}TURNS=200}, \texttt{SDK\_\allowbreak{}MAX\_\allowbreak{}BUDGET\_\allowbreak{}USD=8}, \texttt{SDK\_\allowbreak{}PERMISSION\_\allowbreak{}MODE=accept\allowbreak{}Edits} \\
C2 & claude-opus-4-8 & 2 & 8192 MiB & \texttt{DEEPAGENT\_\allowbreak{}MAX\_\allowbreak{}TURNS=200} \\
C3 & gpt-5.5 & 2 & 4096 MiB & \texttt{CODEX\_\allowbreak{}REASONING\_\allowbreak{}EFFORT=medium} \\
C4 & gpt-5.5 & 2 & 8192 MiB & \texttt{DEEPAGENT\_\allowbreak{}MAX\_\allowbreak{}TURNS=200} \\
C5 & gemini-3.5-flash & 1 & 8192 MiB & \texttt{DEEPAGENT\_\allowbreak{}MAX\_\allowbreak{}TURNS=200} \\
C6 & deepseek/deepseek-v3.2 & 1 & 8192 MiB & \texttt{DEEPAGENT\_\allowbreak{}MAX\_\allowbreak{}TURNS=200}, \texttt{DEEPAGENT\_\allowbreak{}BASE\_\allowbreak{}URL} (OpenRouter), \texttt{DEEPAGENT\_\allowbreak{}EXTRA\_\allowbreak{}BODY} DeepInfra pin \\
\bottomrule
\end{tabularx}
\end{table}

\section{Statistical supplement}

\textbf{Declared tests and multiplicity.} The program plan named four hypotheses. Table~\ref{tab:C-1} lists one primary test per hypothesis and the supplementary tests reported in the paper, with Benjamini–Hochberg q-values computed over the primary family only. H3 for GPT-5.5 is not estimable because both harnesses solve every contest task. The first revision of this paper (also dated 2026-09-08) controlled a seven-test family that mixed primary and supplementary tests, under which the Opus workload interaction had q = 0.10. The family was changed on a reviewer's request to one primary test per hypothesis, and the change is disclosed here because it moves the interaction from q = 0.10 to q = 0.013. The partition itself was chosen after seeing the per-family panel, so the reader should treat the result as post hoc regardless of the family.

\begin{table}[t]
\centering\small
\caption{Declared tests, p-values and q-values}
\label{tab:C-1}
\footnotesize\setlength{\tabcolsep}{4pt}
\sbox{\tblbox}{%
\begin{tabular}{llrr}
\toprule
Hypothesis and test & Role & $p$ & BH $q$ (primary family) \\
\midrule
H1 Opus 4.8: sign-flip on per-task differences & primary & 0.893 & 0.893 \\
H1 GPT-5.5: sign-flip on per-task differences & primary & 0.829 & 0.893 \\
H2: GEE nativeness x vendor & primary & 0.600 & 0.893 \\
H3 Opus 4.8: workload label permutation & primary & 0.003 & 0.013 \\
H1 Opus 4.8: run-pair McNemar (not clustered) & supplementary & 0.868 & n/a \\
H1 Opus 4.8: cluster-adjusted McNemar (Durkalski) & supplementary & 0.782 & n/a \\
H1 GPT-5.5: run-pair McNemar (not clustered) & supplementary & 0.824 & n/a \\
H1 GPT-5.5: cluster-adjusted McNemar (Durkalski) & supplementary & 0.655 & n/a \\
H3 Opus 4.8: stratified sign-flip (null = no effect in either stratum) & supplementary & 0.014 & n/a \\
H3 Opus 4.8: GEE nativeness x contest (logit scale) & supplementary & 0.014 & n/a \\
H3 GPT-5.5: workload label permutation & supplementary & 1.000 & n/a \\
\bottomrule
\end{tabular}}%
\ifdim\wd\tblbox>\linewidth\resizebox{\linewidth}{!}{\usebox{\tblbox}}\else\usebox{\tblbox}\fi

\end{table}

\textbf{Two interaction tests, two nulls.} The stratified sign-flip test flips the sign of each task's paired difference within its stratum. It tests the null that the harness has no effect in either stratum, and a common nonzero effect in both strata would also reject it. The label-permutation test shuffles workload labels across tasks while keeping each task's difference. It tests the null that the workload label carries no information about the effect, which is the interaction null. Both are risk-difference tests on the percentage-point scale. The GEE interaction coefficient is on the log-odds scale and is a different estimand. Their agreement in sign and approximate significance is a consistency check, not one confirmed effect.

\textbf{Cluster-adjusted McNemar.} The run-pair McNemar test in Table~\ref{tab:4-2} pools (task, repeat) pairs and ignores that the two repeats of a task are correlated. The Durkalski et al. (2003) adjustment uses per-task signed discordant counts d\_k = b\_k $-$ c\_k and the statistic T = ($\Sigma$ d\_k)² / $\Sigma$ d\_k², distributed as chi-square with one degree of freedom under the null. It gives p = 0.78 for the Opus contrast and p = 0.65 for GPT-5.5, consistent with the task-bootstrap intervals.

\textbf{pass@k on matched subsets.} Table~\ref{tab:C-2} reports pass@1, pass@2 and the fraction of tasks solved on both repeats on the tasks that have two graded pass/fail repeats in both cells of a contrast (78 tasks for Opus, 77 for GPT-5.5), and the planned-key sensitivity in which any missing or non-pass repeat counts as a failure over all 80 tasks.

\begin{table}[t]
\centering\small
\caption{pass@k on matched task subsets and under the planned-key convention}
\label{tab:C-2}
\footnotesize\setlength{\tabcolsep}{4pt}
\sbox{\tblbox}{%
\begin{tabular}{lrrrrrr}
\toprule
Cell & Matched tasks & pass@1 & pass@2 & both repeats pass & pass@1 (planned keys) & pass@2 (planned keys) \\
\midrule
C1 & 78 & 50.0\% & 57.7\% & 42.3\% & 48.8\% & 56.2\% \\
C2 & 78 & 51.3\% & 57.7\% & 44.9\% & 50.0\% & 56.2\% \\
C3 & 77 & 53.9\% & 61.0\% & 46.8\% & 54.4\% & 62.5\% \\
C4 & 77 & 52.6\% & 57.1\% & 48.1\% & 53.8\% & 58.8\% \\
\bottomrule
\end{tabular}}%
\ifdim\wd\tblbox>\linewidth\resizebox{\linewidth}{!}{\usebox{\tblbox}}\else\usebox{\tblbox}\fi

\end{table}

\textbf{GEE fits.} Table~\ref{tab:C-3} gives the full coefficient tables for the three GEE logistic models (exchangeable within-task correlation, robust standard errors). The OpenAI model's contest coefficient is not estimable because both cells solve every contest task, and its value should be read as a separation artefact.

\begin{table}[t]
\centering\small
\caption{GEE coefficient tables}
\label{tab:C-3}
\footnotesize\setlength{\tabcolsep}{4pt}
\sbox{\tblbox}{%
\begin{tabular}{llrrr}
\toprule
Model & Term & Coefficient (logit) & Robust SE & $p$ \\
\midrule
H2: pass $\sim$ native $\times$ vendor $+$ contest & Intercept & -0.45 & 0.26 & 0.093 \\
H2: pass $\sim$ native $\times$ vendor $+$ contest & C(vendor)[T.openai] & +0.19 & 0.25 & 0.453 \\
H2: pass $\sim$ native $\times$ vendor $+$ contest & native & -0.06 & 0.21 & 0.779 \\
H2: pass $\sim$ native $\times$ vendor $+$ contest & native:C(vendor)[T.openai] & +0.12 & 0.23 & 0.600 \\
H2: pass $\sim$ native $\times$ vendor $+$ contest & contest & +1.92 & 0.37 & 0.000 \\
H3 (Anthropic): pass $\sim$ native $\times$ contest & Intercept & -0.13 & 0.24 & 0.599 \\
H3 (Anthropic): pass $\sim$ native $\times$ contest & native & -0.37 & 0.18 & 0.035 \\
H3 (Anthropic): pass $\sim$ native $\times$ contest & contest & +0.55 & 0.49 & 0.260 \\
H3 (Anthropic): pass $\sim$ native $\times$ contest & native:contest & +1.62 & 0.66 & 0.014 \\
H3 (OpenAI): pass $\sim$ native $\times$ contest & Intercept & -0.40 & 0.25 & 0.104 \\
H3 (OpenAI): pass $\sim$ native $\times$ contest & native & +0.07 & 0.15 & 0.655 \\
H3 (OpenAI): pass $\sim$ native $\times$ contest & contest & +26.96 & 0.34 & 0.000 \\
H3 (OpenAI): pass $\sim$ native $\times$ contest & native:contest & -0.07 & 0.15 & 0.655 \\
\bottomrule
\end{tabular}}%
\ifdim\wd\tblbox>\linewidth\resizebox{\linewidth}{!}{\usebox{\tblbox}}\else\usebox{\tblbox}\fi

\end{table}

\textbf{Per-family panel.} Table~\ref{tab:C-4} gives the per-cell solve rates by task family on the frozen pool. The contest family (lcb) is the stratum in which the Opus pair reverses direction.

\begin{table}[t]
\centering\small
\caption{Per-family panel (frozen pool)}
\label{tab:C-4}
\footnotesize\setlength{\tabcolsep}{4pt}
\begin{tabular}{llrrrrrrr}
\toprule
 & & & \multicolumn{6}{c}{Solve rate (\%)} \\
\cmidrule(lr){4-9}
Family & Kind & Tasks & C1 & C2 & C3 & C4 & C5 & C6 \\
\midrule
AccessCtl & repo & 36 & 39 & 43 & 42 & 36 & 33 & 21 \\
lcb & contest & 19 & 84 & 61 & 100 & 100 & 94 & 40 \\
IdentityApp & repo & 13 & 27 & 42 & 50 & 50 & 46 & 0 \\
FieldSvc & repo & 12 & 46 & 62 & 33 & 42 & 8 & 8 \\
\bottomrule
\end{tabular}

\end{table}

\section{Cost supplement}

\textbf{Original ledger against corrected estimates.} Table~\ref{tab:D-1} gives the per-cell totals the study's ledger recorded during the run, priced by the formula that stood at the time (cache reads subtracted from the input-billed remainder, cache writes not), beside the corrected estimates of \S5.2. The ratio in the last column is the overstatement the two defects produced together. It is 1.14 on C1, where only the cache-write double charge applies, and 4.8 to 7.7 on the deepagents cells, where the double-counted cache tokens were also billed.

\begin{table}[t]
\centering\small
\caption{Original ledger and corrected estimate per cell (USD)}
\label{tab:D-1}
\footnotesize\setlength{\tabcolsep}{4pt}
\begin{tabular}{lrrrrr}
\toprule
Cell & Original ledger (\$) & Corrected estimate (\$) & Original \$ per solve & Corrected \$ per solve & Ledger / corrected \\
\midrule
C1 & 275 & 242 & 3.53 & 3.10 & 1.14$\times$ \\
C2 & 3,129 & 406 & 39.11 & 5.07 & 7.71$\times$ \\
C3 & 251 & 251 & 2.88 & 2.88 & 1.00$\times$ \\
C4 & 1,938 & 292 & 22.54 & 3.39 & 6.64$\times$ \\
C5 & 783 & 164 & 22.37 & 4.67 & 4.79$\times$ \\
C6 & 10 & n/a & n/a & n/a & n/a \\
\bottomrule
\end{tabular}

\end{table}

\begin{figure}[t]
\centering
\includegraphics[width=\linewidth]{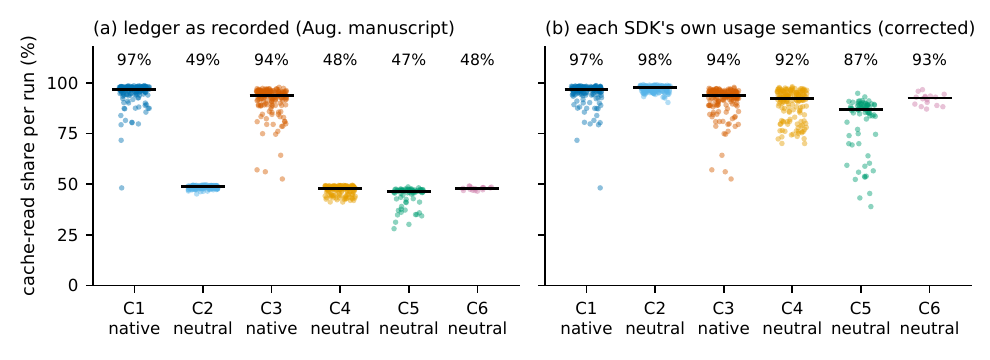}
\caption{Per-run cache-read share by cell, main run. Bars mark medians. Left: as recorded by the ledger, where the host normalizer double-added cache tokens on the neutral harness and produced the ``48\%'' regime the August manuscript reported. Right: the same runs under each SDK's own usage semantics. All six cells cache at 87--98\%.}
\label{fig:cache}
\end{figure}

\textbf{Cache-read share, as recorded and under source semantics.} The figure below shows the per-run cache-read share that the August manuscript presented as a 48\% regime pinned across the deepagents cells, beside the same runs under each SDK's own semantics. The two panels are the same data. The left panel divides cache reads by a total that had those reads counted twice.

\textbf{Decomposition detail.} The component sums of Table~\ref{tab:5-4} use frozen list prices. On Opus 4.8 the native cell has almost no uncached input, because the claude-agent-sdk writes the whole prefix to cache on the first call and reads it on every later call, and its cost is 58\% cache reads, 17\% cache writes and 25\% output. The neutral cell has the same structure with a larger volume of reads (65\%) and output (21\%). On GPT-5.5 both cells spend roughly a quarter on uncached input, half on cache reads and a fifth on output. The C2 minus C1 gap of \$164 is \$124 in cache reads, \$17 in cache writes and \$23 in output. The C4 minus C3 gap of \$41 is \$17 in uncached input, \$19 in cache reads and \$5 in output.

\textbf{Missing-spend allocation.} Table~\ref{tab:5-7} allocates the Anthropic residual under four rules. The rules are illustrative, not estimates. They show the range within which the billed ratio could lie if the residual were attributable to the two cells in those proportions, and they assume that the balance windows and the ledger windows coincide to the day and that the residual is entirely unrecorded model usage.

\section{Probe supplement}

The truncation probe ran on 2026-08-08, after the finish leg, on 12 repository tasks drawn by hash order from the frozen pool (7 AccessCtl, 3 IdentityApp, 2 FieldSvc). The sandbox image was rebuilt with the library stack pinned to the main run's versions, and the only intended change was the runner environment variable that sets deepagents' tool-output truncation to 30,000 bytes. Three unintended changes are recorded in the archive: concurrency 8 rather than 12, one repeat rather than two, and a later date. The probe ledger holds 26 rows for 24 keys (two retries). Table~\ref{tab:5-8} gives the corrected per-run costs, token volumes, cache shares and solve rates on the matched tasks.

The manipulation check that a cap experiment needs is whether the cap ever bound. The canonical events store each tool result as a preview of at most 2,000 characters, and the per-run wire traces store the same preview. Across the 24 main-run sessions on the probe tasks (2,868 tool results under C2 and 2,457 under C4) and the 26 probe sessions (1,333 and 1,331), no stored result is longer than the preview. The archive therefore cannot show how many tool outputs exceeded 30,000 bytes in either arm, and the probe's null is uninterpretable as a test of the truncation hypothesis.

\section{Claims of the August 2026 manuscript and their status}

The August 2026 manuscript circulated internally and was the basis of the first arXiv-ready build. Its claims and their status in this revision are listed below. The full claim-to-evidence table with pointers into the derived data accompanies the replication package.

\begin{table}[t]
\centering\footnotesize\setlength{\tabcolsep}{3.5pt}
\begin{tabularx}{\linewidth}{>{\hsize=1.32\hsize\raggedright\arraybackslash}X>{\hsize=1.32\hsize\raggedright\arraybackslash}X>{\hsize=0.35\hsize\raggedright\arraybackslash}X}
\toprule
August 2026 claim & Status & Where \\
\midrule
A "robust null" for the native-harness premium ($-$1.2 and +1.2 pp) & point estimates unchanged. Re-described as two intervals that do not resolve an average advantage, with the protocol's tests added & \S4.2 \\
Timeout and apply\_error convention shifts cell means by less than 0.5 pp & false. Up to $-$1.25 pp (C3). Both conventions reported & \S4.1 \\
The wall-clock ceiling is a graded failure by design & false. 22 of 81 ceilinged runs passed & \S4.4 \\
A 48\% cache-hit share pinned across deepagents cells, explained as a metric artefact of trajectory shape & an artefact of the telemetry double count. Source-semantics shares are 87–98\% & \S5.1, Appendix E \\
About 4.3$\times$ self-report inflation confined to neutral-harness cells, mechanism unpinned & produced by the double count plus a cache-write pricing error in the study's own pipeline & \S5.1 \\
True cost premium about 2.6$\times$ (Opus) and 1.8$\times$ (GPT-5.5) & withdrawn. Observed-usage ratios 1.3–1.6 and 1.2 with task-bootstrap intervals, and an unresolved billed ordering on Anthropic & \S5.2, \S5.3 \\
11.4M against 3.9M tokens for C2 against C1 on FieldSvc & corrected to 5.69 against 3.91 Mtok & \S5.2 \\
The truncation hypothesis was falsified by the probe & the probe is inconclusive, and the manipulation check is impossible from the archive & \S5.4, Appendix F \\
The design was pre-registered on OSF & a repository-dated plan and a private OSF file deposit of a different design. No registration & \S2.3 \\
Contamination was controlled & bounded to documented cutoffs and private provenance. The canary emission probe was not run & \S2.2 \\
Interim solve rates at the halt were about 5 pp high across cells & higher for the frontier cells and lower for C6 & \S4.1 \\
The ledger held 878 rows & 891 run rows plus two halt records & \S3.2 \\
The sandbox image digest was pinned & \texttt{image\_\allowbreak{}digest: unpinned} on every ledger row. A build fingerprint is recorded on 889 of 891 sessions & Appendix B \\
\bottomrule
\end{tabularx}
\end{table}

\section{Incident history}

The execution incidents summarised in Appendix A are given here in the detail a reader needs to judge whether any could have altered graded data. None did. Scrubbing relabels only ungraded rows, the oracle grades patches independently of the run outcome, and every graded outcome is terminal.

\textbf{Ledger starvation (2026-08-04, dress rehearsal 1).} A Postgres \texttt{numeric} column surfaced as \texttt{decimal.Decimal} in the session row, and \texttt{json.dumps} raised inside every ledger append. The scheduler's worker coroutines died silently, the ledger stayed empty for 28 minutes, and 72 paid sessions ran without the stop-loss seeing them. The fix has two parts. A driver exception becomes a visible, retryable infra row, and an unrecordable append halts the whole run. Both are regression-tested.

\textbf{Stop-loss overshoot (2026-08-04 to 05, dress rehearsal 2).} The accounted stop-loss checked committed spend only after runs finished, so twelve in-flight runs kept spending after the threshold tripped and a €100 cap ended at €275 accounted. The predictive dispatch brake projects committed in-flight spend before each dispatch and throttles ahead of the cap. It held the main run's overshoot to €3,303.34 against a €3,300 cap.

\textbf{Credit exhaustion classified as agent error (2026-08-05, screening).} OpenAI credits ran out mid-screening and 72 consecutive C3 runs failed with a "no credits remaining" message that the taxonomy classified as \texttt{agent\_\allowbreak{}error}, a graded failure. Because those failures reflected the experimenter's account and not the model, a scrub step relabels rows with the credit-failure signature, and rows with the silent-runner-death signature, as retryable infra on the first attempt, and \texttt{--resume} re-runs them as fresh attempt-2 sessions. The scrub recovered 72 C3 and 24 C1 screening rows, 33 rows after the main halt and 38 in the finish leg.

\textbf{Out-of-memory runner deaths (2026-08-05 to 06).} deepagents runners died silently in 4 GiB VMs, and later some claude-agent-sdk runners in 8 GiB VMs. Amendment A5 raised the deepagents cells, and then C1, to 8 GiB. Thirteen C1 deaths at 8 GiB remain unexplained. All were scrubbed and retried and none was graded.

\textbf{OpenRouter default-route probe (2026-08-07, finish leg).} The pre-flight drift gate probed OpenRouter's default route and received 429 responses while the study ran on a DeepInfra-pinned route, blocking resume twice. Amendment A7 pinned the probe to the run's exact backend.

\textbf{Grading disk exhaustion (2026-08-07).} The first full grading pass produced 85 grader errors and 29 timeouts that traced to a full disk on the grading host, not to the oracle. An idempotent purge-and-regrade of the 114 suspect keys produced zero grader errors and the 792 verdicts reported.

\textbf{Telemetry double count (found 2026-09-08).} Described in \S5.1. It affected the accounted currency of the stop-loss and every cost figure of the August manuscript, and no graded verdict.

\section{Reproduction pointers}
\label{app:repro}

\begin{itemize}
  \item \textbf{Token-free rehearsal first.} \texttt{python -m eval.orchestrator run --plan \ldots{} --dry-run} exercises expansion, scheduling, the driver state machine, ledger checkpointing and resume against a fake runtime. \texttt{pytest eval/\allowbreak{}orchestrator/\allowbreak{}tests} runs the invariant families.
  \item \textbf{The ledger is the checkpoint.} Deterministic session ids per (task, cell, repeat, attempt). Graded outcomes are terminal and infra is retried at most twice. An unrecordable append halts the run.
  \item \textbf{Scrub before resume.} Credit-exhaustion and silent-runner-death rows are relabelled retryable infra and re-run as fresh sessions.
  \item \textbf{Pin usage semantics before pricing.} Record per-turn raw usage with its source SDK, declare inclusive or exclusive input per harness, and reconcile ledger totals against the raw events and against billed spend before publishing a dollar figure (\S5).
  \item \textbf{Log exposure to any policy under test.} Per tool call, record the raw output length and whether truncation, summarization or eviction fired (\S5.4).
  \item \textbf{What is not released:} the tasks, golds, hidden tests and canaries. External reproduction of the dataset means re-mining against the published protocol on one's own private code.
\end{itemize}

\bibliographystyle{ACM-Reference-Format}
\bibliography{refs}

\end{document}